\documentclass[letterpaper, 10 pt, conference]{ieeeconf}

\IEEEoverridecommandlockouts

\let\labelindent\relax
\usepackage{amsmath,amssymb,amsfonts,graphicx,booktabs,multirow,algorithm,algorithmic,xcolor,hyperref,siunitx,subcaption,enumitem,mathtools}
\usepackage{dblfloatfix}
\newcommand{\method}{\textsc{VAPS}}

\title{Continue, Abort, or Fall: Viability-Aware Policy Selection (VAPS) for Safe Humanoid Acrobatics}

\author{
  Siwei Ju$^{1,2,*}$,
  Lu Liu$^{3}$,
  Jan Peters$^{1,2,4,5}$,
  Oleg Arenz$^{1,2}$%
\thanks{
This work is part of the Robotics Institute Germany (RIG). This project has been supported by a hardware donation by NVIDIA through the Academic Grant Program.
\newline$^{1}$Department of Computer Science, Technical University of Darmstadt, Germany.
$^{2}$Robotics Institute Germany (RIG).
$^{3}$LimX Dynamics.
$^{4}$Hessian.AI.
$^{5}$German Research Center for AI (DFKI), Research Department: Systems AI for Robot Learning.
\newline$*$ Corresponding author: {\tt\small siwei@robot-learning.de}
}
}

\begin{document}
\IEEEaftertitletext{%
  \vspace{-0.6\baselineskip}%
  \begin{minipage}{\textwidth}%
    \centering%
    \includegraphics[width=0.98\textwidth]{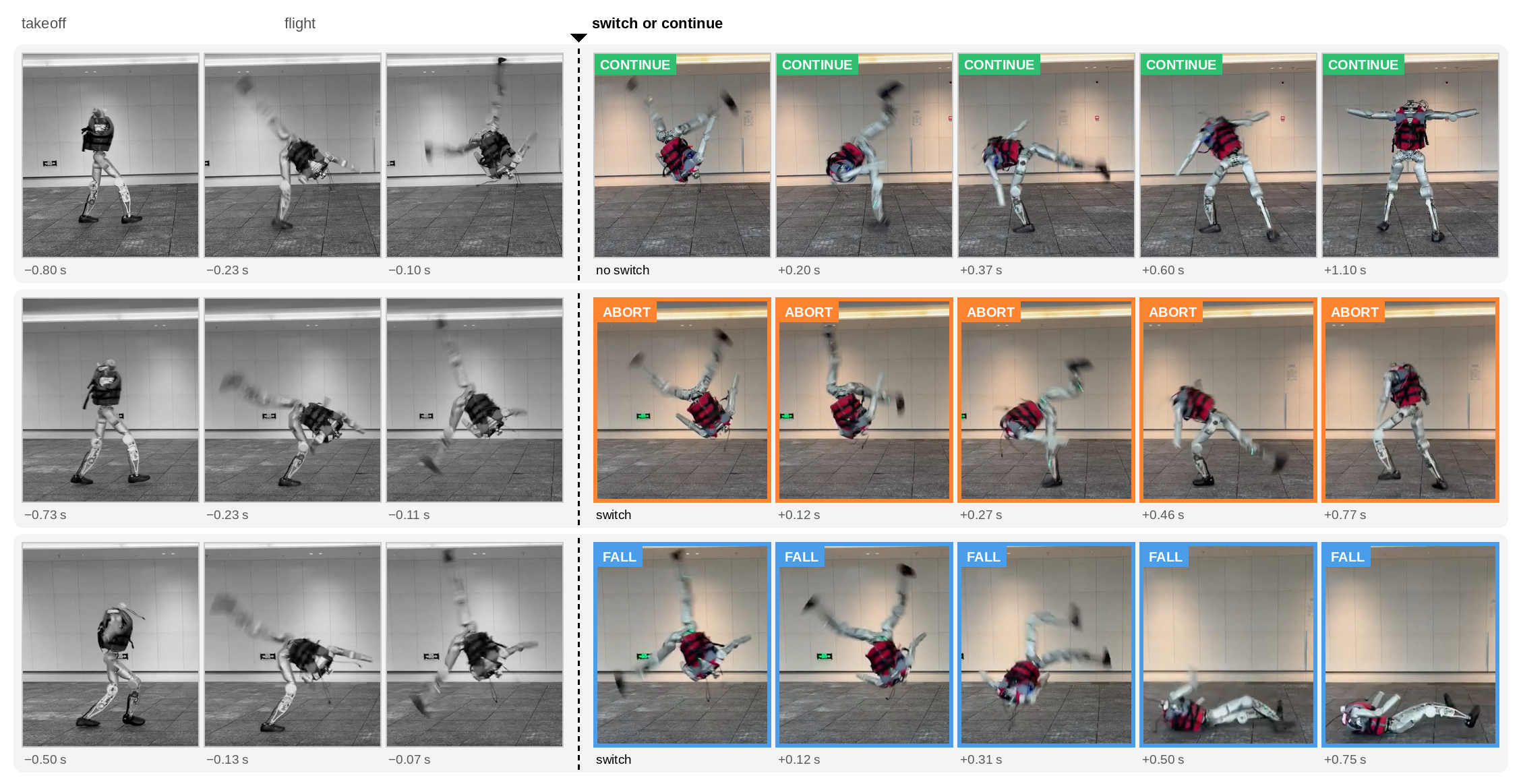}%
    \captionof{figure}{The same side flip on the LimX Oli humanoid with three possible endings. Under the nominal policy the robot completes the flip; when the maneuver is handed mid-flight to the abort policy it lands on its feet, and when it is handed to the protective-fall policy it performs a controlled fall protecting head and hands.}%
    \label{fig:banner}%
  \end{minipage}%
  \vspace{0.8\baselineskip}%
}
\maketitle
\begin{abstract}
Dynamic humanoid motions such as flips risk hardware damage due to suboptimal policies, disturbances or sim-to-real gaps. A motion tracking policy offers no way out once the maneuver leaves its reference, and a backup policy needs to take over to protect the hardware for a minimum-damage landing.
Which backup to use matters as much as when to switch.
We present Viability-Aware Policy Selection (\method{}), which treats safety as a policy-conditioned, receding-horizon decision.
Besides a protective fall policy, we also train an abort policy which can abort the motion at any time, landing on its feet.
At every control step, learned predictors estimate whether the nominal tracking policy and the abort policy remain viable over a short horizon, and a least-sacrificial hierarchy keeps the most task-ambitious behavior that remains viable.
In simulation with randomized disturbances, \method{} sharply reduces head contact and hand contact, which are the dominant sources of hardware damage, with both a Unitree G1 and a LimX Oli; on the LimX Oli, we validate the viability predictors and the full \method{} controller for side-flip motions.
\method{} Pareto-dominates the strongest single-network alternatives we could train, including an end-to-end safe-tracking policy and students distilled from \method{}'s own oracle-routed decisions, in both task success and head impact.
We also show that \method{} is a powerful framework to supervise undertrained policies and protect the hardware. 

Project website: https://vaps-r-al.github.io/
\end{abstract}

\section{Introduction}
\label{sec:introduction}

Humanoid robots are increasingly capable of highly dynamic whole-body behaviors, including running, jumping, and acrobatics~\cite{he_asap_2025, liao_beyondmimic_2025}, but deploying such maneuvers safely remains difficult. Sim-to-real gaps, disturbances or suboptimal policies are common reasons for deployment failure and cause significant hardware damage, especially for highly dynamic acrobatic motions.

Humanoid acrobatics are typically learned as reference-tracking policies~\cite{liao_beyondmimic_2025}. A tracking policy is rewarded for staying close to its reference, and its training episodes terminate as soon as it strays from it; at deployment, once the maneuver has left the reference, a backup must take over. Most existing safety mechanisms nonetheless cast this intervention as a binary choice between continuing the nominal controller and activating a single backup policy or safety shield, which either steers the robot back to a safe region~\cite{thananjeyan_recovery_2021, alshiekh_safe_2017, yang_safe_2022} or executes a safe fall~\cite{meng_safefall_2025}. Highly dynamic acrobatics are fast and short, lasting only one to two seconds. When something goes wrong, there is little time left to correct or recover, only to abort safely. 
Depending on the severity of the deviation, a robot that has left its reference may still be able to abort the motion with a controlled landing on its feet rather than a fall, whereas even a learned safe fall can cause serious damage at these speeds. 
A single switch is therefore too coarse, particularly in the intermediate regime where the maneuver is already lost but a controlled landing is not. 
Folding execution and recovery into one policy~\cite{lei_kung_2026, xu_unified_2025} does not restore the missing granularity either.

We address these challenges with \emph{policy-conditioned, receding-horizon viability}: whether a state is safe depends on which closed-loop behavior runs next, and should be judged over a short, continuously re-evaluated window rather than by a single guess for the end of the episode. We instantiate this as Viability-Aware Policy Selection (\method{}), which selects among the \textit{nominal} tracking policy and backups ordered by decreasing task ambition: an \textit{abort} policy that seeks a controlled feet-first landing, and a \textit{protective-fall} policy that minimizes harm once upright recovery is no longer viable. A policy-dependent predictor, which encodes a short observation history, re-estimates each behavior's short-horizon viability at every control step, and \method{} preserves the most ambitious behavior that remains viable. As the predictor judges the state and the candidate behavior rather than the cause of risk, the same mechanism recovers a mature policy from disturbances, supervises a new and unproven one, and handles a stop request issued by an operator or a workspace monitor. As hardware failures are expensive to collect, all failure cases are generated in simulation, and each state is reloaded under every candidate behavior.

We summarize our main contributions as follows:
\begin{itemize}[leftmargin=*]
    \item \emph{Least-sacrificial safety.} \method{} formulates intervention during a dynamic maneuver as a selection over an ordered hierarchy of behaviors: continue, abort, and protective fall, committing to the smallest sacrifice of task ambition that remains viable.

    \item \emph{Policy-conditioned, receding-horizon viability.} We learn viability predictors for the nominal policy and the abort policy from counterfactual rollouts and evaluate them continuously from proprioceptive history.

    \item \emph{Analysis of single-policy alternatives.} We investigate monolithic alternatives, including end-to-end RL policies and students distilled from \method{} with DAgger, and show that \method{} dominates all of them.

    \item \emph{Hardware validation.} We develop and test \method{} in simulation on Unitree G1 and LimX Oli, deploy it and validate the sim-to-real transfer on LimX Oli.
\end{itemize}

\section{Related Work}
\label{sec:related_work}

A safety architecture for a dynamic maneuver must supply four things: a controller that executes the motion, behaviors that protect the robot when it fails, a signal that says when to intervene, and a structure that organizes the intervention. We review the literature on each in turn and then position \method{} against all four.

\textbf{Humanoid motion tracking and acrobatics.} Reference-conditioned reinforcement learning lets humanoids reproduce complex motions from human demonstrations, from phase-conditioned tracking~\cite {peng_deepmimic_2018} to large-scale retargeting, privileged learning, and sim-to-real transfer~\cite {he_omnih2o_2024, he_asap_2025, liao_beyondmimic_2025}, and now covers jumping, spinning, cartwheels, and parkour~\cite {liao_beyondmimic_2025, zhuang_humanoid_2024, wu_perceptive_2026, xie_kungfubot_2025}. These methods enable the motion, but don't tell what the robot should do once the reference is infeasible at deployment.

\textbf{Protective and recovery behaviors.} Fall protection spans model-based contact selection that minimizes impact impulse~\cite {kumar_learning_2017}, learned protective policies triggered by a continuously running fall predictor~\cite {meng_safefall_2025}, curriculum-discovered protective postures~\cite {shi_discovering_2025}, and robot-agnostic stylized falling~\cite {strauch_robot_2025}. These supply the behavior a robot needs once the maneuver is lost, and supply it well. What none of them does is a controlled landing while nominal motion is not possible any more, but a fall is still avoidable. 

\textbf{Deciding when to intervene.} The trigger has been cast as a learned shield or constraint critic~\cite {thananjeyan_recovery_2021, alshiekh_safe_2017}, as control barrier functions and reachability~~\cite {ames_control_2017, bansal_hamilton-jacobi_2017}, as a model-predictive safety filter~\cite {wabersich_predictive_2021}, as Hamilton--Jacobi values scaled to humanoids by a geometrically distributed receding horizon~\cite {chen_-reachability_2026}, and, closest to humanoid falls, as a continuously running fall predictor~\cite {meng_safefall_2025}. Each computes one safety value for one policy. A quantity of that form cannot identify the severity of the lost maneuver and whether a controlled abort is still possible.

\textbf{Organizing the intervention.} Given a signal, the common structure is binary handover from the task policy to a recovery policy, whether triggered by a constraint critic~\cite {thananjeyan_recovery_2021}, by predicted violations on a legged robot~\cite {yang_safe_2022}, or by a reach-avoid backup in a hierarchical controller~\cite {he_agile_2024}; these designs assume one nominal objective, one generic recovery behavior, and a mature nominal policy~\cite {liu_safe_2023}. Closest to our setting, concurrent work on humanoid locomotion learns a task-agnostic stoppability value and, on a commanded stop, chooses between holding upright and a damping fall~\cite {long_safestop_2026}, while the controlled abort is more challenging to train for aerobatic motions than locomotion. A growing line removes the structure instead, blending execution and recovery in one policy: martial-arts motions, flips included, that recover from disturbed or fallen states~\cite {lei_kung_2026}; fall prevention, impact mitigation, and standing recovery unified~\cite {xu_unified_2025} and distilled into a vision-based student~\cite {azulay_vigor_2026}; or locomotion learned jointly with recovery~\cite {ren_stubborn_2026, lu_unified_2026}. 

\begin{figure*}[t]
    \centering
    \includegraphics[width=.82\textwidth]{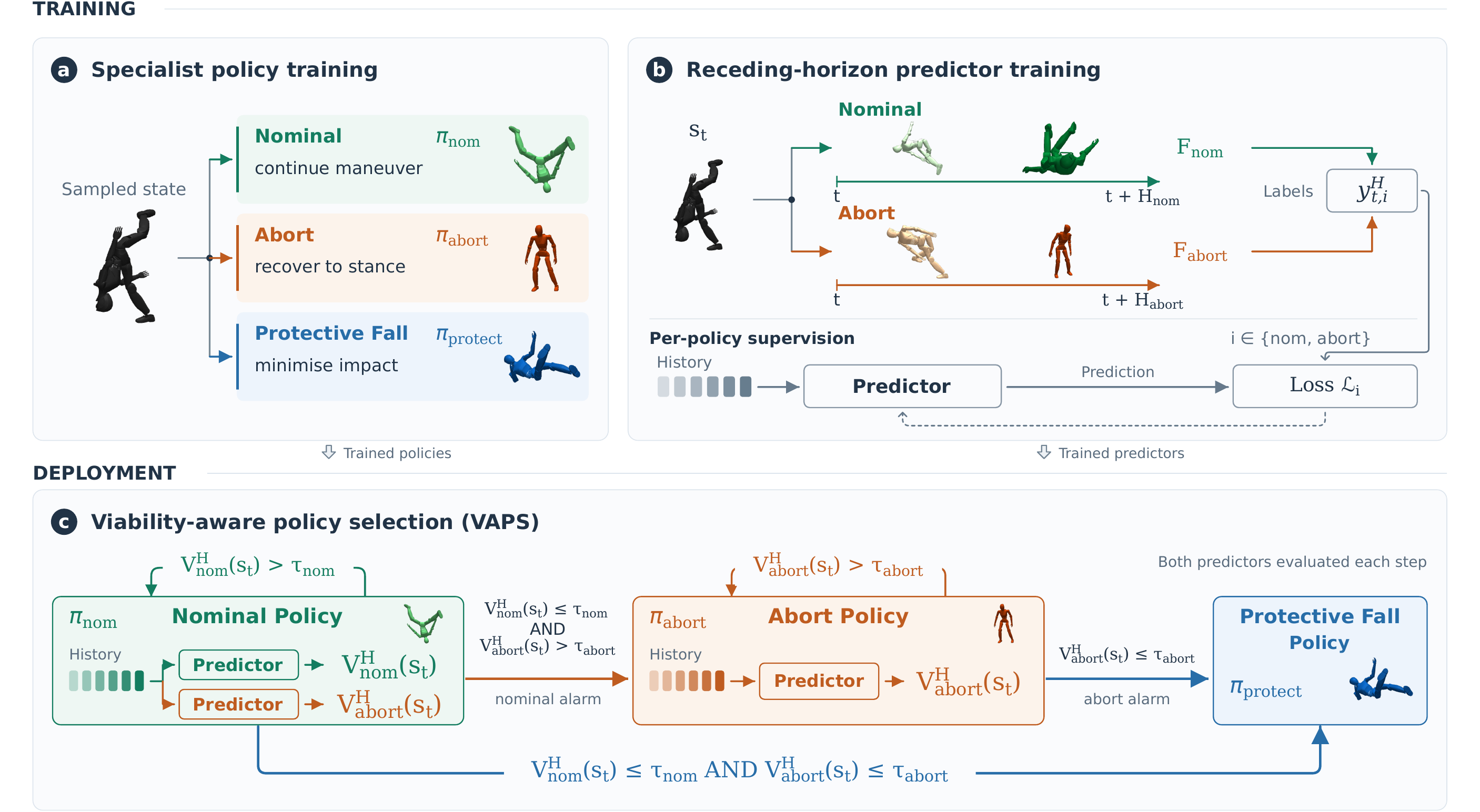}
    \caption{Overview of \method{}. (a) Specialist policy training. (b) Receding-horizon predictor training. (c) \method{}: The least-sacrificial rule of policy selection.}
    \label{fig:overview}
\end{figure*}

\textbf{Positioning.} \method{} takes the motion tracking policy as given and fixed. It is whichever tracking controller that is deployed, mature or not. For backup behaviors, it trains two backup specialists rather than one: an abort policy that aims for a controlled landing, and a protective-fall policy as the last resort. In terms of the predictive trigger, where prior work computes one safety value for one policy, a per-policy predictor re-estimates each candidate behavior's finite-horizon viability at every control step. Last but not least, for the intervention, prior work uses a switch which hands one policy over to one backup, \method{} keeps several named behaviors (continue, abort, fall) and retains the most ambitious one that is still viable. We also investigate the single-policy alternatives to \method{}.

\section{Method}
\label{sec:method}

\method{} has three parts, illustrated in Fig.~\ref{fig:overview}. A set of specialist policies covers the maneuver and the two ways out of it (\S\ref{sec:specialist_policies}). For the nominal and the abort policy, corresponding viability predictors estimate whether they can still meet their own objectives over a finite horizon (\S\ref{sec:policy_conditioned_prediction}). A least-sacrificial selection rule reads these estimates at every control step and retains the most task-ambitious behavior that remains viable (\S\ref{sec:policy_selection}). In addition, \S\ref{sec:monolithic_alternatives} investigates single-network alternatives to \method{}.

\subsection{Specialist Policies}
\label{sec:specialist_policies}
Humanoid acrobatic motions are learned by imitating reference motions captured from humans and retargeted to each robot \cite{liao_beyondmimic_2025}. To keep the policy on its reference, every training episode is terminated as soon as the robot deviates from it. The resulting policy has never seen a departure from the reference, so it cannot save the robot once one occurs at deployment: a backup policy has to take over.

The states at which control is handed over differ in how dangerous they are, and different backup behaviors from the same state lead to different chances of damage. We therefore define two backup behaviors with conflicting objectives: an abort policy that seeks a controlled, feet-first landing, and a protective-fall policy that limits harm once a fall has become unavoidable. Together with the nominal controller, this decomposes the safe execution of an acrobatic motion into three specialist behaviors with separate objectives: motion success, controlled abort, and protective falling. The corresponding policies are
\begin{equation}
    \Pi = \{\pi_{\mathrm{nom}}, \pi_{\mathrm{abort}}, \pi_{\mathrm{protect}}\},
\end{equation}
labeled Nominal, Abort, and ProtFall in tables and figures.

\textbf{Nominal Policy.} $\pi_{\mathrm{nom}}$ denotes whichever motion-tracking policy is deployed or under evaluation. In our experiments, it tracks a single reference clip, obtained by motion capture and retargeting, under
\begin{equation}
    r_{\mathrm{nom}} = r_{\mathrm{track}} - c_{\mathrm{reg}},
\end{equation}
where $r_{\mathrm{track}}$ is the tracking reward in terms of reference joint configuration, body poses, and body velocities, and $c_{\mathrm{reg}}$ penalizes action rate, soft joint-limit violations, and self-collision. The structure of the maneuver enters only implicitly, through the reference clip. 

\textbf{Abort Policy.} $\pi_{\mathrm{abort}}$ minimizes damage by seeking a controlled landing: an upright base with only the feet in ground contact. It is trained to reach and hold this configuration, without reference to the original maneuver, under
\begin{equation}
    r_{\mathrm{abort}} = r_{\mathrm{alive}} + r_{\mathrm{upright}} + r_{\mathrm{support}} - c_{\mathrm{reg}},
\end{equation}
where $r_{\mathrm{alive}}$ rewards every step that does not end in a hard termination, $r_{\mathrm{upright}}$ rewards keeping the base's vertical axis upright, and $r_{\mathrm{support}}$ rewards simultaneous contact of both feet. Training resets are drawn from a precomputed dataset of \emph{switch states}: simulator states snapshotted while rolling out the frozen nominal policy offline under domain-randomized disturbances, weighted toward harder-to-recover states in the flight and landing phases. The abort policy is thus trained on the handoff distribution of the very nominal policy it backs up.

\textbf{Protective-Fall Policy.} A controlled landing is not always possible. When a fall is unavoidable, $\pi_{\mathrm{protect}}$ takes over to reduce its severity. Two parts of a humanoid need particular protection: the head, which carries cameras and other easily broken sensors, and the hands, which are dexterous, fragile, and expensive. While shielding these, the robot should fall so as to minimize contact impacts and the joint loads they induce. $\pi_{\mathrm{protect}}$ is trained from the same switch states, concentrated around flight and landing, under
\begin{equation}
    r_{\mathrm{protect}} = r_{\mathrm{alive}} - c_{\mathrm{head}} -
    c_{\mathrm{hand}} - c_{\mathrm{joint}} - c_{\mathrm{torque}},
\end{equation}
where $c_{\mathrm{head}}$ and $c_{\mathrm{hand}}$ penalize contact forces on the head and hand regions beyond the robot's static support load, and $c_{\mathrm{joint}}$ and $c_{\mathrm{torque}}$ penalize reaction-force and torque overload relative to per-joint-group thresholds. Rollouts end only at timeout or on head-region contact, so success for $\pi_{\mathrm{protect}}$ means mitigated harm rather than task success.

\subsection{Policy-Conditioned Receding-Horizon Viability}
\label{sec:policy_conditioned_prediction}
After introducing the specialists, we need the trigger signal which can be used to switch between them. For the nominal and the abort policy, we estimate how likely it is that running $\pi_i$ ($i \in \{\text{nom}, \text{abort}\}$) from the current state stays free of failure over the next $H$ control steps, and we re-estimate this at every control step.

\textbf{Viability.} Let $F(s)$ be a failure predicate on states, and let $\Delta_i(s_t)$ be the number of control steps until $F$ first fires when $\pi_i$ is run from $s_t$ ($\Delta_i = \infty$ if it never fires). The $H$-step viability of $\pi_i$ is the probability that no failure occurs within $H$ steps,
\begin{equation}
\label{eq:policy_viability}
    V_i^H(s_t) = \Pr\left( \Delta_i(s_t) > H \,\middle|\, s_t, \pi_i \right).
\end{equation}
We define $F$ as a \emph{physical fall}: terrain contact by any non-foot collision geometry. We deliberately do not reuse the tracking-error terminations that shape $\pi_{\mathrm{nom}}$. They fire on deviations which the robot often survives and end the episode before any fall is observable. The same $F$ serves both policies.

\textbf{Rollout and label.} Each policy gets its own dataset $\mathcal{D}_i$ of simulated rollouts under domain randomization and injected disturbances. Crucially, rollouts start from the states at which that policy is actually entered at deployment. Matching training states to the states the predictor is queried on affected the switching decision more than architecture, history length, and sensing combined. Every step $t$ of a rollout, with its realized time to failure, yields the viability label $y_{t,i}^{H} = \mathbb{I}\left[ \Delta_i(s_t) > H \right]$.

Due to partial observability, the predictor needs the $N$ most recent observations available to $\pi_i$: $\widehat{V}_{i,t} = f_{\theta_i}\!\left(s_{t-N+1:t}\right) \in (0,1)$, 
where $f_{\theta_i}$ ends in a scalar sigmoid and $\widehat{V}_{i,t}$ estimates $V_i^H(s_t)$.

\textbf{Discounted viability estimation.}
For each state $s_t$, we place a geometrically decaying distribution over future offsets,
\begin{equation}
    p_\gamma(k) = \gamma^k.
\end{equation}
We define the corresponding discounted data and predictor distributions over the future offset $k$ and viability outcome $y\in\{0,1\}$ as
\begin{align}
    p_{\mathrm{data}}^\gamma(y\mid s_t)
    &=
    \sum_{k=0}^{H}
    p_\gamma(k)\,
    p_{\mathrm{data}}(y_{t+k}=y\mid s_t),
    \\
    p_{\theta_i}^\gamma(y\mid s_t)
    &=
    \sum_{k=0}^{H}
    p_\gamma(k)\,
    p_{\theta_i}(y_{t+k}=y\mid s_t).
\end{align}

We train the predictor by minimizing the cross-entropy between these discounted distributions,
\begin{equation}
    \mathcal{L}_i
    =
    \mathbb{E}_{s_t \sim \mathcal{D}_i}
    \left[
        \mathcal{H}\!\left(
            p_{\mathrm{data}}^{\gamma}(y \mid s_t),
            p_{\theta_i}^{\gamma}(y \mid s_t)
        \right)
    \right].
\end{equation}
Ignoring terms independent of $\theta_i$, this is equivalent to
\begin{equation}
\begin{aligned}
    \mathcal{L}_{i,t}
    = -\sum_{k=0}^{\infty}\gamma^k
    \Big[&
        p_{\mathrm{data}}(y_{t+k}=1\mid s_t)
        \log \widehat V_{i,t+k} \\
        &+
        p_{\mathrm{data}}(y_{t+k}=0\mid s_t)
        \log(1-\widehat V_{i,t+k})
    \Big].
\end{aligned}
\end{equation}
For $\gamma<1$, prediction errors closer to the current state receive greater weight than those farther into the future.

\subsection{Viability-Aware Policy Selection}
\label{sec:policy_selection}
The three policies have increasing sacrifice. $\pi_{\mathrm{nom}}$ completes the maneuver. $\pi_{\mathrm{abort}}$ gives up the maneuver but keeps the robot upright. $\pi_{\mathrm{protect}}$ gives up staying upright and only limits the damage of a fall that can no longer be avoided. Each step down the ladder costs more of the task and exposes the hardware to more risk. The robot should therefore descend only as far as the predicted viability forces it to. At deployment, the executed policy is the least sacrificial one that is still predicted viable,
\begin{equation}
\label{eq:hierarchical_policy_selection}
    \pi_t =
    \begin{cases}
        \pi_{\mathrm{nom}},
        & \widehat{V}_{\mathrm{nom},t} > \tau_{\mathrm{nom}},\\[4pt]
        \pi_{\mathrm{abort}},
        & \widehat{V}_{\mathrm{nom},t} \leq \tau_{\mathrm{nom}}
          \;\land\; \widehat{V}_{\mathrm{abort},t} > \tau_{\mathrm{abort}},\\[4pt]
        \pi_{\mathrm{protect}},
        & \text{otherwise}.
    \end{cases}
\end{equation}
Eq.~\eqref{eq:hierarchical_policy_selection} is evaluated at every control step subject to the current level of the hierarchy, and transitions are irreversible within one maneuver. So an abort that starts to fail is itself escalated to the protective fall; we call this the \emph{full hierarchy}. A \emph{one-shot} variant consults the abort predictor once at the nominal alarm and then commits to the chosen backup.

Both thresholds are selected empirically. $\tau_{\mathrm{nom}}$ trades false alarms against lead time. A higher threshold hands off earlier, giving the abort more time to recover, but it also interrupts more maneuvers that would have succeeded. We require a false-alarm rate (FAR) of at most $1\%$ on held-out stable episodes. In our experiments, this FAR budget corresponds to $\tau_{nom}=0.2$, $\tau_{abort}=0.87$.

The rule responds to predicted viability, not to the cause of its loss, and therefore the same hierarchy protects a mature policy against external disturbances, and it protects the hardware while a new, untested policy is being deployed. We investigate this aspect further in §\ref{sec:untrusted_results}.

\subsection{Monolithic Alternatives}
\label{sec:monolithic_alternatives}

Three policies, two predictors, and a switching rule are a lot of machinery, so it is natural to ask whether it is possible to replace them with one policy. We take this possibility seriously. On the same robot, motion, and reward terms, we build the two strongest single-network alternatives.

\textbf{Single Network RL.} The most direct route is to fold safety
into the task reward,
\begin{equation}
    r_{\mathrm{e2e}} = r_{\mathrm{nom}} - \lambda\,(c_{\mathrm{head}} + c_{\mathrm{hand}} + c_{\mathrm{joint}} + c_{\mathrm{torque}}).
\end{equation}
This policy optimizes the same damage terms as $\pi_{\mathrm{protect}}$. Because Abort and ProtFall have conflicting objectives, we can only merge one of them into this policy. The only thing it lacks is the ability to switch objectives: it must decide, inside one set of weights, when tracking is still worth the risk. To let it learn that decision, we remove the tracking-error terminations of $\pi_{\mathrm{nom}}$, so an episode that loses the maneuver runs on into the fall and learns how to do it. The weight $\lambda$ is then the only knob between task and damage, and this trade-off is known to be fragile in constrained RL~\cite{achiam_constrained_2017, stooke_responsive_2020}. 

\textbf{Distilled Student of \method{} and DAgger.} We distill a single student from a teacher that is at least as good as \method{} itself: an oracle-routed pipeline that, at every alarmed state, runs counterfactual rollouts under each specialist and takes the behavior that succeeds. The student therefore imitates perfect policy selection. Each student is a single feedforward network on the nominal policy's observation~\cite{lee_learning_2020}, trained by supervised learning on a stratified mix of benign, pre-alarm, and post-switch segments. Faithful imitation has a blind spot. Because post-switch samples constitute only a small fraction of the data, we investigate to oversamle them, and use DAgger ~\cite{ross_reduction_2011} to add student-visited states that the oracle routes to the protective fall. Both make the student more protective.

We also want to understand if memory is what separates the two designs. The predictors read an observation history, so a single-frame student could be missing information rather than structure. To rule this out, we also train distillation students with temporal context, using either a stacked window of past observations or a recurrent (GRU) encoder, on the same data and teacher.

\section{Experiments}
\label{sec:experiments}

The experiments address five questions:
\begin{enumerate}[leftmargin=*,nosep]
  \item How do the individual specialists perform, in simulation and on hardware? (\S\ref{sec:viability_results})
  \item How accurately is finite-horizon viability predicted relative to prior fall-prediction baselines, and does that accuracy hold on the real robot? (\S\ref{sec:prediction_results})
  \item How does \method{} perform in simulation and on hardware: how much safety is gained, and at what cost in task success? (\S\ref{sec:pipeline_results})
  \item How does \method{} compare with monolithic alternatives, obtained either by end-to-end reinforcement learning or by distilling \method{} into a single policy? (\S\ref{sec:structure_results})
  \item Can \method{} supervise partially trained policies it was not built around? (\S\ref{sec:untrusted_results}) 
\end{enumerate} 
Simulation results in this section use the Unitree G1 performing a front flip trained in MJLab at \SI{50}{Hz}; results for the LimX Oli are similar and are briefly reported in Table \ref{tab:main_results}; we also report transfer experiments on the physical LimX Oli humanoid at the same rate with a side flip motion. Unless stated otherwise, results are aggregated over five seeds.

\subsection{Specialist Policies}
\label{sec:viability_results}

Table~\ref{tab:specialist_viability} rolls the three policies out at three domain-randomization (DR) scales. The two backups are rolled out from identical handoff states. At the training scale all three succeed without touching the ground. At the intermediate scale the nominal begins to lose the maneuver, while the protective fall avoids head contact on essentially every state it is handed and the abort returns in 89\% to a stance. At the strongest scale, the policies differ markably in performance: the nominal completes only 15\% and puts its head on the ground at 62\%, the abort still returns in 62\% to a stance, and protective falling is the most reliable on its own criterion (96\%). The maneuver thus becomes unrecoverable long before safe recovery does, and the backups are preferable over different bands of severity.

\begin{table}[t]
    \centering
    \caption{Specialist performance across domain randomization (DR) scales. Specialist success: Nominal -- motion success, Abort -- upright stance, ProtFall -- head clearance.}
    \label{tab:specialist_viability}

    \begin{tabular}{llccc}
        \toprule
        Policy & DR & Specialist Success & Head contact & Hand contact \\
        \midrule
        \multirow{3}{*}{Nominal}
        & 1 & $100.0\pm0.0\%$ & $0.0\pm0.0\%$ & $0.0\pm0.0\%$ \\
        & 2 & $86.2\pm1.6\%$ & $8.1\pm1.1\%$ & $13.6\pm1.6\%$ \\
        & 4 & $15.0\pm2.2\%$ & $62.0\pm2.9\%$ & $82.3\pm2.0\%$ \\
        \midrule
        \multirow{3}{*}{Abort}
        & 1 & $100.0\pm0.0\%$ & $0.0\pm0.0\%$ & $5.0\pm0.0\%$ \\
        & 2 & $89.0\pm8.2\%$ & $9.0\pm6.5\%$ & $11.0\pm7.4\%$ \\
        & 4 & $62.0\pm13.0\%$ & $17.0\pm8.4\%$ & $26.0\pm6.5\%$ \\
        \midrule
        \multirow{3}{*}{ProtFall}
        & 1 & $100.0\pm0.0\%$ & $0.0\pm0.0\%$ & $4.0\pm2.2\%$ \\
        & 2 & $98.0\pm2.7\%$ & $2.0\pm2.7\%$ & $9.0\pm8.9\%$ \\
        & 4 & $96.0\pm4.2\%$ & $4.0\pm4.2\%$ & $22.0\pm12.5\%$ \\
        \bottomrule
    \end{tabular}
\end{table}

\subsection{Viability Prediction}
\label{sec:prediction_results}

We evaluate the nominal predictor on held-out rollouts spanning all DR scales, at the frozen validation threshold of \S\ref{sec:policy_selection}, against SafeFall-style final-outcome predictors \cite{meng_safefall_2025}, which estimate whether the episode ends in a fall (in our notation, this corresponds to $H=\infty, \gamma=1$), retrained on our rollouts with the same splits and the same $\leq 1\%$ false-alarm rule. At the same episode-level false-alarm rate, \method{} detects more falls and warns about $30 - 60\%$ earlier (Table~\ref{tab:safefall_comparison}).

For \cite{meng_safefall_2025}, pushing its decision boundary out toward our horizon buys lead time only up to $t_2=\SI{500}{ms}$,\footnote{$t_2$ is the baseline's labelling window: a state is labelled as falling if it lies within $t_2$ of the end of a failed episode. Larger $t_2$ asks the baseline to fire earlier.} after which lead time regresses together with detection, because the baseline's label rule caps how much of the falling phase can be supervised: a final-outcome label can fire only once the outcome is settled, often past the point where an upright landing is still reachable, and this prediction also depends on the executed policy. 

\textbf{Hyperparameters for the predictor.} For \method{} predictors, we swept for $\gamma \in \{\boldsymbol{0.90}, 0.95, 0.97, 1.00\}$, $H \in \{10, 25, \boldsymbol{50}\}$; \textbf{GRU}, Transformer and MLP encoders, $N \in \{10, \boldsymbol{20}, 50\}$; five seeds per cell, and selected the one with the best combination of lead time and detection accuracy. 

\begin{table}[t]
    \centering
    \caption{Predictor comparison against SafeFall-style final-outcome baselines. FAR: false alarm rate. LT: lead time.}
    \label{tab:safefall_comparison}

    \resizebox{\linewidth}{!}{
    \begin{tabular}{lcccc}
        \toprule
        Method & Episode Acc. $\uparrow$ & Detection $\uparrow$ & FAR $\downarrow$ & LT mean $\uparrow$ \\
        \midrule
        SafeFall ($t_2=\SI{100}{ms}$) & $95.5\pm2.1\%$ & $85.8\pm7.0\%$ & $0.4\pm0.8\%$ & $0.205\pm0.012$\,s \\
        SafeFall ($t_2=\SI{200}{ms}$) & $95.0\pm1.5\%$ & $83.3\pm5.1\%$ & $\mathbf{0.0\pm0.0}\%$ & $0.259\pm0.014$\,s \\
        SafeFall ($t_2=\SI{500}{ms}$) & $95.0\pm0.0\%$ & $84.2\pm1.9\%$ & $0.4\pm0.8\%$ & $0.316\pm0.063$\,s \\
        SafeFall ($t_2=\SI{1000}{ms}$) & $93.5\pm1.6\%$ & $79.2\pm5.1\%$ & $0.4\pm0.8\%$ & $0.281\pm0.046$\,s \\
        \method{} ($\gamma=0.9,H=50$) & $\mathbf{97.5\pm1.5}\%$ & $\mathbf{91.7\pm5.1}\%$ & $\mathbf{0.0\pm0.0}\%$ & $\mathbf{0.416\pm0.042}$\,s \\
        \bottomrule
    \end{tabular}}
\end{table}

\begin{figure}[t]
    \centering
        \includegraphics[width=0.85\linewidth]{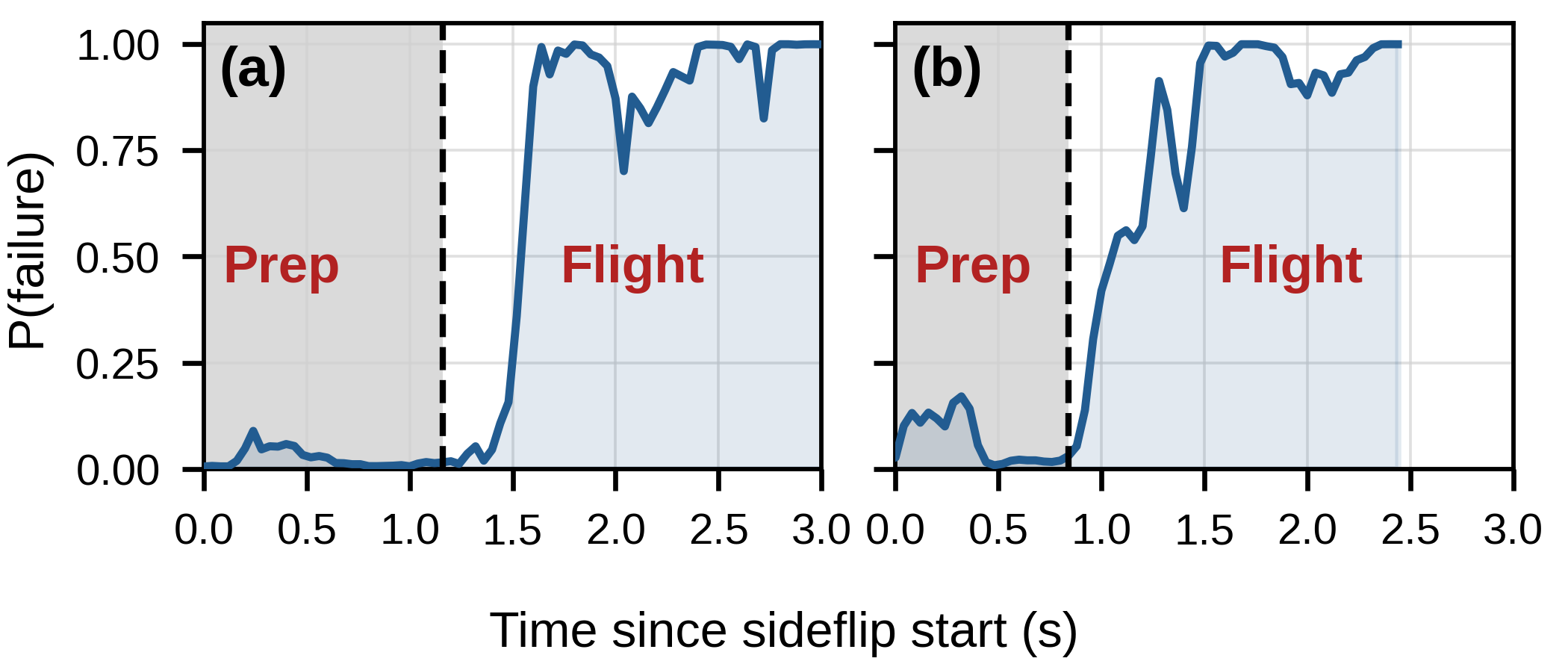}
    \caption{Predictor output for two representative failed real-robot side flips. The dashed line marks take-off.}
    \label{fig:real_robot_prediction}
\end{figure}

\textbf{Real-robot evaluation.} On 13 side-flip trials with the LimX Oli shown in Fig. \ref{fig:banner} -- eight successful executions and five failures -- the deployed predictor classifies every trial correctly with no false alarm, firing on each failure before the corresponding simulated termination with a mean lead of \SI{0.544}{s}, consistent with simulation (Fig.~\ref{fig:real_robot_prediction}). Thirteen trials bound the false-alarm rate only loosely, but they establish that the quantity the selector depends on survives the sim-to-real transfer.

\begin{table*}[t]
    \centering
    \caption[Closed-loop performance under randomized disturbances]{Closed-loop performance under randomized disturbances. $\pm$ is over five seeds. Results are on Unitree G1 except for the last two rows.
    }
    \label{tab:main_results}

    \resizebox{\linewidth}{!}{
    \begin{tabular}{lccccc|cc}
        \toprule
        & \multicolumn{5}{c|}{All episodes (\%)}
        & \multicolumn{2}{c}{Fallen episodes only, median\,[mean]} \\
        \cmidrule(lr){2-6}\cmidrule(lr){7-8}
        Method
        & Success $\uparrow$ & Head $\downarrow$ & Hand $\downarrow$ & Other $\downarrow$ & Fall $\downarrow$
        & Pk.\ head (N) & Pk.\ non-foot (N) \\
        \midrule
        G1, Nominal only
        & \textbf{62.1}\,$\pm$\,0.2 & 27.7\,$\pm$\,0.2 & 36.6\,$\pm$\,0.2 & 36.3\,$\pm$\,0.1 & 37.9\,$\pm$\,0.2
        & 540.2\,$\pm$\,13.8\,[954.8\,$\pm$\,122.0] & 2124.6\,$\pm$\,33.9\,[2608.6\,$\pm$\,352.2] \\
        \midrule
        \multicolumn{8}{l}{G1, \method{} --- \emph{shared alarm rule, $\tau_{\mathrm{nom}}=0.20$}} \\
        \method{} (Nominal--Abort)
        & 59.6\,$\pm$\,0.4 & 5.0\,$\pm$\,0.4 & \textbf{7.4}\,$\pm$\,0.5 & \textbf{5.5}\,$\pm$\,0.3 & \textbf{7.4}\,$\pm$\,0.5
        & 833.5\,$\pm$\,139.1\,[1076.0\,$\pm$\,72.1] & 2219.5\,$\pm$\,87.2\,[2110.4\,$\pm$\,189.7] \\
        \method{} (Nominal--ProtFall)
        & 59.6\,$\pm$\,0.4 & 2.0\,$\pm$\,0.1 & 11.8\,$\pm$\,0.6 & 39.9\,$\pm$\,0.5 & 40.4\,$\pm$\,0.4
        & 0.0\,$\pm$\,0.0\,[21.8\,$\pm$\,2.6] & 1871.8\,$\pm$\,14.2\,[2058.6\,$\pm$\,17.4] \\
        \method{} (one-shot)
        & 59.6\,$\pm$\,0.4 & 2.2\,$\pm$\,0.2 & 9.3\,$\pm$\,0.4 & 25.6\,$\pm$\,1.7 & 26.2\,$\pm$\,1.7
        & 0.0\,$\pm$\,0.0\,[62.2\,$\pm$\,12.9] & 1855.3\,$\pm$\,38.1\,[2073.4\,$\pm$\,77.2] \\
        \method{} (full hierarchy)
        & 59.6\,$\pm$\,0.4 & \textbf{1.6}\,$\pm$\,0.2 & 8.8\,$\pm$\,0.5 & 25.6\,$\pm$\,1.8 & 26.1\,$\pm$\,1.7
        & 0.0\,$\pm$\,0.0\,[28.8\,$\pm$\,3.8] & \textbf{1835.7}\,$\pm$\,38.4\,[2032.3\,$\pm$\,20.9] \\
        \midrule
        \multicolumn{8}{l}{G1, \emph{Monolithic, memoryless}} \\
        End-to-end RL ($\lambda{=}10$)
        & 61.9\,$\pm$\,0.9 & 24.0\,$\pm$\,1.6 & 36.1\,$\pm$\,1.3 & 36.7\,$\pm$\,0.8 & 37.9\,$\pm$\,1.0
        & 388.2\,$\pm$\,95.0\,[846.7\,$\pm$\,82.4] & 2051.5\,$\pm$\,34.5\,[2372.4\,$\pm$\,69.5] \\
        Distilled (base)
        & 57.4\,$\pm$\,0.8 & 23.1\,$\pm$\,1.1 & 33.6\,$\pm$\,1.1 & 30.4\,$\pm$\,0.8 & 34.0\,$\pm$\,1.1
        & 756.3\,$\pm$\,50.7\,[1100.8\,$\pm$\,59.6] & 2453.8\,$\pm$\,33.6\,[2608.8\,$\pm$\,77.4] \\
        Distilled (+\,protect)
        & 58.0\,$\pm$\,0.6 & 17.9\,$\pm$\,1.4 & 36.7\,$\pm$\,0.5 & 36.1\,$\pm$\,0.5 & 38.4\,$\pm$\,0.6
        & 18.2\,$\pm$\,40.8\,[712.8\,$\pm$\,65.2] & 2329.8\,$\pm$\,22.9\,[2552.1\,$\pm$\,22.9] \\
        Distilled (+\,protect, DAgger)
        & 56.3\,$\pm$\,0.7 & 10.1\,$\pm$\,1.3 & 39.7\,$\pm$\,1.0 & 40.5\,$\pm$\,0.7 & 42.3\,$\pm$\,1.0
        & 0.0\,$\pm$\,0.0\,[326.6\,$\pm$\,69.3] & 2132.8\,$\pm$\,57.2\,[2456.4\,$\pm$\,54.8] \\
        \midrule
        \multicolumn{8}{l}{G1, \emph{Monolithic with memory}} \\
        Distilled frame-stack 5 (base) 
        & 55.5\,$\pm$\,3.7 & 25.4\,$\pm$\,0.3 & 39.2\,$\pm$\,3.3 & 35.0\,$\pm$\,1.2 & 39.7\,$\pm$\,3.3
        & 675.0\,$\pm$\,129.2\,[1039.2\,$\pm$\,85.0] & 2383.7\,$\pm$\,137.6\,[2498.0\,$\pm$\,192.7] \\
        Distilled GRU (base)
        & 58.7\,$\pm$\,1.1 & 24.2\,$\pm$\,0.8 & 34.8\,$\pm$\,1.1 & 32.2\,$\pm$\,0.8 & 35.2\,$\pm$\,1.1
        & 774.2\,$\pm$\,49.9\,[1121.0\,$\pm$\,41.9] & 2482.6\,$\pm$\,42.5\,[2619.2\,$\pm$\,52.5] \\
        Distilled GRU (+\,protect)
        & 56.0\,$\pm$\,1.0 & 15.4\,$\pm$\,0.9 & 35.3\,$\pm$\,1.4 & 34.3\,$\pm$\,1.9 & 37.0\,$\pm$\,1.3
        & 0.0\,$\pm$\,0.0\,[646.8\,$\pm$\,55.6] & 2256.1\,$\pm$\,44.9\,[2502.4\,$\pm$\,33.8] \\
        Distilled GRU (+\,protect, DAgger)
        & 56.1\,$\pm$\,1.3 & 17.9\,$\pm$\,1.2 & 33.6\,$\pm$\,1.4 & 32.2\,$\pm$\,1.3 & 34.8\,$\pm$\,1.5
        & 227.2\,$\pm$\,204.1\,[835.9\,$\pm$\,101.8] & 2369.2\,$\pm$\,61.4\,[2574.4\,$\pm$\,92.8] \\

        \midrule
        \multicolumn{8}{l}{\emph{LimX Oli, side flip}} \\
        Nominal only (LimX Oli)
        & 80.7\,$\pm$\,0.3 & 15.5\,$\pm$\,0.2 & 19.3\,$\pm$\,0.3 & 15.5\,$\pm$\,0.2 & 19.3\,$\pm$\,0.3
        & 1308.5\,$\pm$\,17.3\,[1277.1\,$\pm$\,15.7] & 4561.2\,$\pm$\,61.4\,[4560.6\,$\pm$\,30.3] \\
        \method{} (full hierarchy, LimX Oli)
        & 79.2\,$\pm$\,0.9 & 1.1\,$\pm$\,0.2 & 4.8\,$\pm$\,1.4 & 7.0\,$\pm$\,2.1 & 9.1\,$\pm$\,2.2
        & 0.0\,$\pm$\,0.0\,[143.0\,$\pm$\,41.6] & 2960.9\,$\pm$\,323.7\,[2912.2\,$\pm$\,356.0] \\

        \bottomrule
    \end{tabular}}
    \vspace{2pt}
\end{table*}

\subsection{Full Pipeline Evaluation}
\label{sec:pipeline_results}

The upper block of Table~\ref{tab:main_results} compares \method{} with two single-backup ablations obtained by restricting Eq.~\eqref{eq:hierarchical_policy_selection} to one backup: Nominal--Abort routes every alarm to the abort, Nominal--ProtFall to the protective fall. All four rows share the same alarm rule, so they leave the nominal on the same episodes and motion success is identical, $59.6\%$ against $62.1\%$ for the nominal policy. \method{} costs $2.5$ points; the remaining columns show what each backup does with the alarmed episodes.

The two ablations separate the roles of the backups: the abort protects the whole body, while the protective fall keeps the head off the ground and reduces the impact. Nominal--Abort has the lowest fall rate in the table, but an abort that fails arrives in whatever attitude it lost control in, so its falls are the most damaging and its head contact ($5.0\%$) the highest of any switching row. Nominal--ProtFall protects the head but is committed to the ground by construction, so it falls more often than the nominal policy.

\method{} retains the benefits of both by selecting and switching between them. One-shot \method{} sends each alarm to the abort if the abort viability predictor judges the state recoverable and to the protective fall otherwise: head contact stays at $2.2\%$, within $0.2$ points of Nominal--ProtFall, while the fall rate drops by $14.2$ points to $26.2\%$, consistently across all five seeds. A full-hierarchy \method{} keeps the abort predictor running during the abort, and escalates to the protective fall when its prediction drops. It further improves head-protection: head contact falls to $1.6\%$, below the protective-fall ablation, and the mean peak head force among fallen episodes more than halves, with fall rate and motion success unchanged. The full hierarchy also records the lowest median peak non-foot force in the table.

At the handoff step, where the policy and both predictors evaluate, the three networks take on average \SI{0.29}{ms} on one CPU core, \SI{1.45}{\percent} of the control period.

\textbf{Hardware experiments.} In practice, we would like to test each of the backup policies in terms of its own success rate. We conducted the individual policy tests on hardware by triggering the switch manually on the LimX Oli: handed the maneuver mid-motion, the abort returned the robot to a stance in $17$ of $19$ trials, while the protective fall policy managed to protect the head and hands in $12/16$ trials. 

For the full-hierarchy \method{} hardware evaluation, we adjusted the nominal-predictor threshold to \(\tau_{\mathrm{nom}}=0.3\) to induce more frequent handoffs to the backup policies. Across 18 trials, the controller executed four routing patterns: nominal only (\(2/2\) successful), nominal \(\rightarrow\) abort (\(9/10\)), nominal \(\rightarrow\) abort \(\rightarrow\) protective fall (\(3/4\)), and nominal \(\rightarrow\) protective fall (\(1/2\)). Overall, the selected specialist achieved its intended objective in \(15/18\) trials (\(83.3\%\)), including trials with externally introduced disturbances.

\subsection{VAPS versus Monolithic Policies}
\label{sec:structure_results}
\begin{figure}[t]
    \centering

    \includegraphics[width=\linewidth]{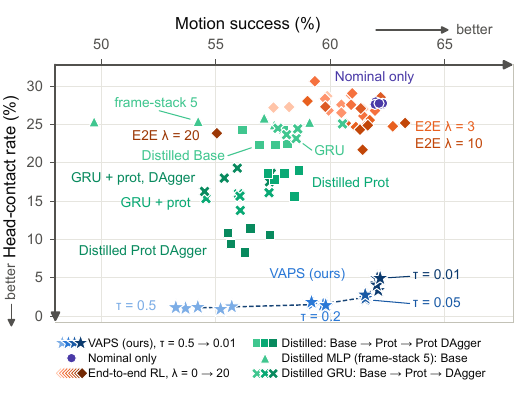}

    \caption{Head-contact rate against motion success on the Unitree G1 results of Table~\ref{tab:main_results}; down and right is better.}
    \label{fig:pareto_front}
\end{figure}
In this section, we turn to the second part of Table~\ref{tab:main_results} and to Fig.~\ref{fig:pareto_front} to ask whether the structure of \method{}, with its specialist policies and viability predictors, can be replaced by a single policy: an end-to-end RL policy\footnote{Trained from scratch, the policy either shortcuts the flip without inverting or refuses it altogether, so we warm-start it from the frozen $\pi_{\mathrm{nom}}$ at a reduced learning rate. We sweep $\lambda \in \{0, 0.3, 1, 3, 6, 10, 20, 30\}$ with five seeds each and compare against the best operating point.}, or one of several students distilled from \method{}, as introduced in §\ref{sec:monolithic_alternatives}\footnote{Protective falling makes up only $12.7\%$ of the post-switch supervision, so the student rarely sees how to fall. Two further arms deliberately tilt it toward protection. \emph{+\,protect} draws half of its post-switch frames from protective-fall segments, and \emph{+\,DAgger} adds student-visited states that the oracle routes to the protective fall.}.

Fig.~\ref{fig:pareto_front} places every Unitree G1 row of Table~\ref{tab:main_results} on the plane of motion success against head-contact rate, one marker per seed, and the three families occupy three distinct regions of it. \method{} runs along the bottom edge, where its threshold traces a frontier from about \SI{55}{\percent} success at \SI{2}{\percent} head contact to \SI{62}{\percent} at \SI{4.6}{\percent}.
The end-to-end policies never leave the nominal's corner: they lower head contact by at most four points, before the weight breaks training and success scatters. The distilled students descend in head-contact rate, giving up success but removing up to half of the base student's head contact, and then stuck at about \SI{10}{\percent}. 

Different architectures of the distilled \method{} students improve the protection, but sacrifice the motion success. A five-frame stack is not better than a single frame, and a recurrent state helps only a little. The greatest improvement is from the training data. The DAgger shard, whose segments begin at the alarm, takes the memoryless student from \SI{17.9}{\percent} to \SI{10.1}{\percent} head contact yet leaves the recurrent student almost where it was, because of the missing information for the hidden states. The students have partially learned how to protect the head, but they are missing the trigger to determine when they should switch to the protection behavior.

As a result, \method{} dominates on the Pareto frontier. At $\tau=0.01$, \method{} matches the nominal's success, while achieving substantially lower head-contact rates than the best end-to-end and distilled alternatives; every higher $\tau$ trades success for still less head contact, down to \SI{1.1}{\percent} at $\tau=0.5$. Up to seed spread on the success axis, the non-dominated set of Fig.~\ref{fig:pareto_front} consists of \method{} points only. Folding the specialists into one network preserves the skills but loses the decision of how much of the task to give up, which is exactly the part the explicit structure keeps inspectable.

\subsection{Supervising New Policies}
\label{sec:untrusted_results}

\begin{figure}[t]
    \centering
    \includegraphics[width=\columnwidth]{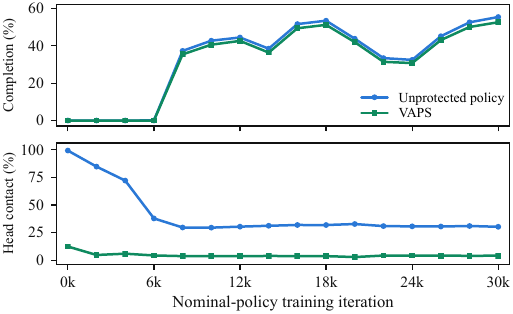}
    \caption{\method{} applied to $16$ checkpoints of an independently trained nominal policy. }
    \label{fig:untrusted_policy}
\end{figure}

Because the abort policy, the protective-fall policy and the abort predictor depend only on the handoff states and not on the nominal policy's weights, supervising a new nominal policy requires retraining only the nominal predictor. \method{} can therefore protect the hardware during the early deployment of a new policy or a new robot. To test this on suboptimal, undertrained policies, we train the nominal policy again from scratch, save a checkpoint every $2{,}000$ iterations, and deploy \method{} on each of the $16$ checkpoints. Fig. \ref{fig:untrusted_policy} shows that \method{} protects the head from damage while keeping nominal success across all checkpoints. 

\section{Conclusion and Outlook}
\label{sec:conclusion}
Humanoid robots are acquiring dynamic skills faster than they are acquiring safe ways to deploy them, and the cost is hardware damage during development and deployment. This work starts from a simple observation: whether a state is safe cannot be decided from the state alone. It depends on what the robot does next and on how much time remains to save it. \method{} builds on this by judging each available behavior on its own terms, over a short receding horizon, and keeping the most capable behavior that is still predicted to be viable. Between continuing the maneuver and falling, it places a third option that prior work usually omits: aborting the maneuver while staying upright. We validate \method{} in simulation and on a physical humanoid performing side flips.

The structure also proved hard to replace. Single networks trained on the same objectives, and students distilled from \method{} itself, learn how to fall but not when to give up the task, and they fail to match its combination of task success and safety. Keeping the backup policies and their predictors separate is also what makes \method{} portable. The same construction protects a mature controller against disturbances, supervises a partially trained one, and serves a stop requested from outside the robot, across two humanoid platforms. Looking ahead, extending viability prediction to other tasks is a natural next step. More broadly, deciding when to abandon a task, and how to do so safely, is a question that reaches well beyond acrobatics.

\bibliographystyle{IEEEtran}
\bibliography{main}

\end{document}